# MRGazer: Decoding Eye Gaze Points from Functional Magnetic Resonance Imaging in Individual Space

Xiuwen Wu, Rongjie Hu, Jie Liang, Yanming Wang, Bensheng Qiu, Xiaoxiao Wang

*Abstract*—Eye-tracking research has proven valuable in understanding numerous cognitive functions. Recently, Frey et al. provided an exciting deep learning method for learning eye movements from fMRI data. However, it needed to co-register fMRI into standard space to obtain eyeballs masks, and thus required additional templates and was time consuming. To resolve this issue, in this paper, we propose a framework named MRGazer for predicting eye gaze points from fMRI in individual space. The MRGazer consisted of eyeballs extraction module and a residual network-based eye gaze prediction. Compared to the previous method, the proposed framework skips the fMRI co-registration step, simplifies the processing protocol and achieves end-to-end eye gaze regression. The proposed method achieved superior performance in a variety of eye movement tasks than the co-registration-based method, and delivered objective results within a shorter time (~ 0.02 Seconds for each volume) than prior method (~0.3 Seconds for each volume).

*Index Terms*—Eye tracking, deep convolutional neural networks, object detection, digital morphology operation

## I. INTRODUCTION

Eye is the window of mind, and eye movement is relevant to cognitive processes in visual perception, attention, reading, and memory [1]. Therefore, the pattern of eye movement becomes a critical research indicator to help reveal the brain function of normal people and reveal biomarkers of psychiatric disorders [2], [3]. fMRI is a widely used noninvasive neuroimaging method that may detect the neural activity of the whole brain. However, eye tracking is merely used in fMRI experiments because MR-compatible eye trackers are expensive and require for comprehensive setups and calibrations [8].

However, these methods both rely on co-register the fMRI with MR-template or well-designed eye masks [16] to extract eyeballs. These lead to some problems: 1. The process is time consuming and requires a manually quality check, which make the protocol complicated, especially for real-time fMRI-based eye tracking; 2. The choice of atlas template is complicated and determined by age, sex, and disease specificity [17], [18]. 3. The unexpected failures in this process might induce a waste of data. To overcome the abovementioned problems, we propose an end-to-end eye movement regression pipeline based on fMRI images in individual space without co-registration. The overview of the pipeline is shown in Fig. 1(a). First, the fMRI images in individual space will be collected by the designed eye movement scans trials before the MRI protocol. The eye movement scans involves that a series of dots are fully distributed on the screen, and the participants are asked to fixate on these points during scans. Therefore, the fMRI image at a certain point in time and the corresponding gaze point coordinates will support model training. Second, the eyeballs signals will be extracted by the 3D object detection from fMRI images in individual space. The architecture we used is the Retina-Net which consists of a 3D Res-Net for feature extraction and a feature pymarid network (FPN) for multi-features fusion. The trained Retina-Net can obtain the eyeballs from fMRI images directly. In order to get the bounding box for Retina-Net training, we considered the relationship of connected components between the brain and eyes and used the digital morphology open operator algorithm to split the eyeballs and brain area (Fig. 1(b). The process can find the connected components of eyeballs and we chose the bounding box to guarantee the whole connected components of eyes can be contained in it. And then, the bounding box can be used to train the model for extracting eye signals automatically. Third, the eye signals will be fed into a 3D Res-Net for decoding the eye gaze points. The architecture of the regression model is shown in Fig. 1(c).

Our contributions are summarized as follows:
1) Using digital morphology, we obtained a large number of fMRI eye masks as a feature.
2) The pipeline does not require T1 images or templates for co-registration.
3) The approach achieves superior performance in multiple experiments.
4) We trained a classifier to detect outliers to remove potential samples that would report poor results before the features were fed into prediction networks.
5) We propose a reasonable and practicable end-to-end paradigm to solve eye fixation tasks with real-time attributes.

Xiuwen Wu and Rongjie Hu contributed equally.
Correspongding autors:Bensheng Qiu;Xiaoxiao Wang.
XiuwenWu, Rongjie Hu, Jie Liang, Yanming Wang, Bensheng Qiu and Xiaoxiao Wang are with the Center for Biomedical Imaging, University of Science and Technology of China, HeFei 230026, China(e-mail: wuxiuwen@mail.ustc.edu.cn,rongjiehu@mail.ustc.edu.cn,jie_liang@mail.ustc.edu.cn,ming1258@ustc.edu.cn,bqiu@ustc.edu.cn,wang506@ustc.edu.cn).

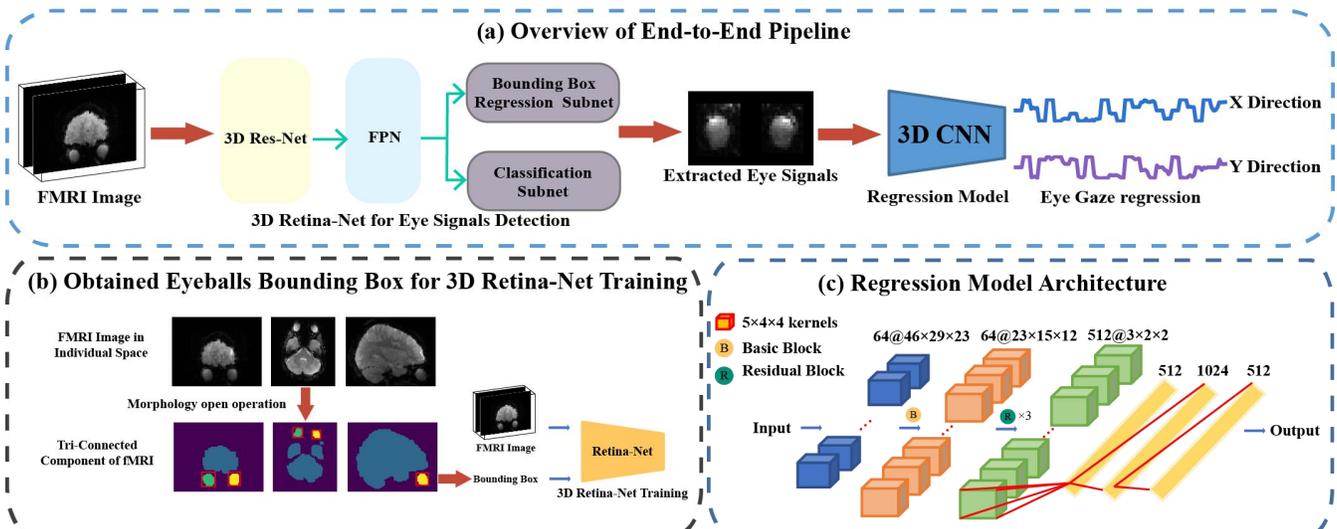

Fig. 1. Illustration of the proposed MRGazer. (a) The overview of end-to-end pipeline for eye gaze regression. The fMRI images in individual space were fed into 3D Retina-Net for eye signals detection. The extracted eye signals were then decoded by 3D Res-Net for eye gaze prediction in x and y directions separately. (b) Obtained eyeballs bounding box for 3D Retina-Net training. The morphology open operation can find the connected components of eyes to help generate the bounding box, which was used for 3D Retina-Net training. (c) 3D Res-Net for eye gaze regression. The features of eyeballs were fed into a Res-Net.

## II. RELATED WORKS

### A. FMRI-BASED EYE TRACKING METHOD USING SVR

The PEER approach estimates the direction of gaze in each fMRI volume by the SVR method [9]. It begins with the collection of an additional eye gaze calibration fMRI scanning, during which each subject was introduced to fixate on the dots on the screen. The fMRI data were then preprocessed, and an eye mask was employed to isolate the signal around the eyeballs. Then, the isolated signal from the calibration scan was fed into the PEER to train a subject-specific model, which was used for fMRI-based eye tracking in futural fMRI scans of the same subject. [11] evaluated the PEER's performance during naturalistic viewing and found that a 1.5 min long calibration scan was enough for training the model. A few works [8], [9], [11] have employed the method due to its low accuracy [12] and lack of across-subject generality.

### B. NATURAL IMAGE-BASED EYE TRACKING USING DEEP LEARNING

With the development of deep learning, many convolutional neural network (CNN)-based models have been proposed to solve cameraless eye regression on natural images [13], [14], [15]. Krafka et al. [13] created the first large-scale dataset for natural image eye tracking. Then, they proposed a model named iTracker, which considered head pose relative to the camera and the pose of eyes relative to the head. The model achieved a significant reduction in error on the hardware of mobile phones and tablets. Zhang et al. [14] presented the MPIIGaze dataset for real-world conditions. They also proposed a CNN-based architecture that achieves state-of-the-art performance on the current datasets. Zhang et al. [15] achieved an improvement of up to 14.3% on MPIIGaze and 27.7% on EYEDIAP for person-independent 3D gaze estimation. These models all take into account the positional relationship between the head and the eyeballs because under real-world conditions, these factors are unconstrained. The difference is that in an MRI scan, the subject's head is fixed inside the coil and kept still. This makes the position of the head have very little effect on the performance of the models.

### C. FMRI-BASED EYE TRACKING METHOD USING DEEP LEARNING

To the best of our knowledge, Frey, Nau and Doeller [16] proposed the DeepMReye method and were the first to employ the CNN model for fMRI-based eye tracking. Firstly. They created eye masks, including the adjacent optic nerve and muscle area, manually in a standard structural template, and then they created a group average functional template by averaging the coregistered functional images of 29 participants. To ensure that the eye masks contained the eyeballs, all the images were processed by three coregistration steps, and each voxel underwent two normalization steps. Finally, the coregistered and normalized voxels were taken as the input of the deep learning model. After feature extraction, Res-Net was used for the prediction of eye movement. The 'DeepMReye' model is a good start for a fMRI-based eye tracking method using deep learning. It allows for camera-less gaze monitoring at a moderate temporal resolution even in many existing datasets and when the eyes are closed. It also provides a powerful tool for cognitive state readout during scanning.

## III. MATERIALS AND METHODS

### A. DATA DESCRIPTION

#### 1) FIXATION DATA

These data were downloaded from the Healthy Brain Networks (HBN) Biobank (http://fcon_1000.projects.nitic.org/indi/cmi_healthy_brain_network/), of which the participants ranged from 5 to 21 years old[9]. The MRI data collected at the Rutger site (from release 1.0 to release 8.0) were employed in the study. During the fixation and saccade task, the participants were introduced to fixate at a fixation target that moved through 27 locations within a window of 14.2×.9 degrees visual angle, and each fixation lasted for 4 seconds. The task was scanned repeatedly three times (named peer1, peer2, and peer3 in the HBN). After removing the data with visible artifacts and violent head motion (≥3 mm), we eventually obtained MRI data of peer1 (N=702), peer2 (N=399), and peer3 (N=398). Note that the eye movements were not tracked during the MRI scan, the participants were assumed to follow the instructions, and the positions of the fixation targets served as labels for training and testing. Considering that every fixation began with an eye movement of saccade, the first frame of each gaze point would include saccade-related information and was removed from further processing.

### 2) NATURALISTIC VIEWING DATA

These data were also from the Rutger site (from release 1.0 to release 8.0) of the Healthy Brain Networks (HBN) Biobank. During the naturalistic viewing, the participants were introduced to view four animated films. We employed the data from viewing the animation "The present" and eventually included 207 participants after removing the data with visible artifacts and violent head motion(≥3 mm). The eye movements were not monitored during the MRI scan. In the EEG protocol of naturalistic stimuli, eye movement information was monitored by an eye tracker (iView-X Red-m, SensoMotoric Instruments GmbH) at a sampling rate of 120 Hz.

### 3) PRO-SACCADE AND ANTI-SACCADE DATA

The pro-saccade and anti-saccade tasks are widely used eye movement paradigms, which help to investigate brain function and diagnose various neurological or psychiatric disorders [19], [20], [21]. During the task, the participant was introduced to saccade to the target (pro-saccade task) or the opposite direction of the target relative to the center of the screen (anti-saccade task) [21], [22]. The data were downloaded from two datasets in OpenNeuro: the OpenNeuro Dataset ds000120 'visual guide saccade' (both pro-saccade and anti-saccade data [23] and the OpenNeuro Dataset ds000119 'incentive anti-saccade' (anti-saccade data [22]. The two datasets were both scanned in a 3.0-T Siemens Allegra scanner at the Brain Imaging Research Center, University of Pittsburgh, Pittsburgh, PA, and eye movements were monitored simultaneously with a long-range optics eye-tracking system (Model R-LRO6, Applied Science Laboratories, Bedford, MA). The result of the tasks, rather than the eye tracking data, were provided in the datasets: a saccadic event marked with 'correct' means a saccade to the target in pro-saccade or to the opposition of the target in anti-saccade; a saccadic event marked with 'incorrect' means no saccade or saccade to the wrong direction. Thus, the position of the 'correct' target in the pro-saccade task served as training labels, and the visual field contralateral to the position of the 'correct' target in the anti-saccade task served as testing labels. Eventually, 69 participants with data in the 'visual guide saccade' with pro- and anti-saccades and 24 participants with data in the 'incentive anti-saccade' with anti-saccades were included.

### B. MATHEMATICAL MORPHOLOGY TO EXTRACT EYE SIGNALS

Mathematical morphology provides a set of operators to process digital imaging [24], [25], [26], [27]. The approach is efficient and obtains exact results in different fields of digital image processing, such as noise reduction, shape recognition, feature extraction, and edge detection [28], [29], [30]. Two factors affect the result, structural element and morphology operators, which constitute the essence of the algorithm. The structural element is defined manually before operators apply it to the image, and it determines the shape of the extracted information from the image. Morphological operators are used to express how objects or shapes in images interact with structural elements.

Ideally, the human brain and eyes can be thought of as three connected areas in fMRI because of the high BOLD signal. However, in reality, the nerves between the eyes and brain generate signals strong enough to make the three connected areas form a holistic one. Meanwhile, some tissues and artifacts add complexity to the 'triple-link area', as shown in Fig. 12(a). Therefore, the brain and eyes cannot meet the triplet-connected region hypothesis. However, according to the mathematical morphology approach, eyeball masks are still available. Before using the open operator, the fMRI image needs to be binarized. Taking into account the interference of redundant information, a rough region of interest (ROI) containing eye features and part of the brain is obtained by reducing the scale of the z-axis and using the fMRI image mean value as the threshold to binarize the image. Because of low eye signals in some images, the voxel values of the eyes are lost in the binarization process. Hence, the binarization threshold using the mean value of the ROI requires a coefficient $\gamma$ to adjust. The function of binarization can be expressed as follows:

$$y = \begin{cases} 1 & x \geq T_R * \gamma \\ 0 & x \leq T_R * \gamma \end{cases} \quad (1)$$

where x is the value of the voxel, y is the binarized output, $T_R$ is the mean of the ROI, and $\gamma$ is the coefficient. After that, the structural element is defined as a cube shape at $3 \times 3 \times 3$ slides in the image and performs the open operation. This process breaks the tiny connected region between the brain and eyeballs and removes extraneous connected regions formed by noise. Then, the process counts the number of connected regions in the image and calculates the volume of each. Finally, the region with the largest volume is deleted, and eyeball regions are filtered out, as shown in Fig. 2(b).

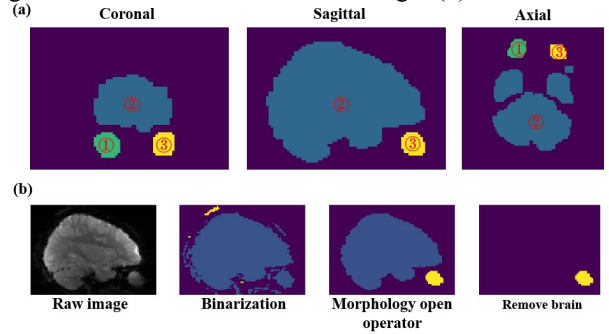

Fig. 2. The process of eye signal extraction. (a) Ideally, because of high signals, the eyes and brain form three connection areas. The figure

shows the coronal, sagittal, and axial sections. The numbers 1 and 2 are the areas of the eyes, and 3 represents the areas of the brain. (b) First, some nerve tissues around the eyes have high signals in fMRI imaging, and we binarize fMRI images to not always obtain three areas that can make a distinction between the brain and eyes because of the tiny connections. Additionally, there are some noise areas that pervade the whole image. Thus, the mathematical morphology open operator helps to break the tinny connection and strip eyes from the brain connection area. Then, we can easily remove brain areas based on the volume of connected regions. Finally, we can obtain the eye area and generate bounding boxes to package the eyes.

### C. Deep learning architecture and training parameters

#### 1) Regression model

Researchers have proposed convolutional neural network (CNN) models to handle eye gaze regression tasks in natural images [13], [14], [15]. These CNN-based models all considered the position of the face in the image and the relative positional relationship between eyes and face as well as which face and eyes both input into the networks. However, during fMRI scans, the subject's head is fixed in the electromagnetic coil. This means that the position of each participant's head and eyes have slight differences compared to the natural image. [8], [11], [16], [31] demonstrated that signals of eyes extracted from fMRI provide full information to predict eye fixations. Consequently, we only fed eyes into networks rather than both signals of the eyes and head. We choose 12 layers of the residual network (Res-Net) proposed by [32] as our deep learning architecture balancing the performance and complexity of networks (Fig. 1(e)). A series of 3D residual blocks with batch normalization (BN) layers and convolutional layers constitute our architecture. The shape of the input feature is 48×30×24. In the first layer, we used 5×4×4 convolutional filters; then, the two $3 \times 3 \times 3$ convolutional filters constituted the basic block and followed the three residual blocks with 3×3×3 convolutional filters. Finally, three fully connected layers with 512, 2048, and 512 units are used for regression. The structure of the outlier detection model is the same as that of the regression model.

For the training regression model, we selected Adam as the optimizer algorithm, in which the optimal parameter values for the learning rate were set to 5×10-4. The batch size was set to 128, and the trick of early stopping was performed to stop the training after the 60th epoch. The model is trained and predicted separately for the x and y directions. The optimal parameters of the Res-Net models were obtained by minimizing the mean square error loss function fitting with regression tasks. The loss function is defined as (2):

$$MSE = \frac{1}{n}\sum_{i=1}^{n}(y_i - \hat{y}_i)^2 \quad (2)$$

where n is the size of the batch, $y_i$ is the real gaze position and $\hat{y}_i$ is the predicted results. The MSE of x and y directions will be calculated separately.

The outlier detection model is the same as the regression model, except for the fully connected layers. The learning rate was set at $1\times10^{-4}$; the batch size was set at 128, and the model was trained for 15 epochs. The best results of the model trained in the fivefold validation were selected to classify anomalies.

The Peer3 and naturalistic viewing data of eye signals ground truth were used to train the model. The learning rate was set to $1\times10^{-4}$, the batch size was 32, and the model was trained for 120 epochs. The loss function was the focal loss, which enables to detection of small objects from a vast background. The number of anchors was 78, which confirms the coverage of different volumes of eyeballs.

#### 2) Eye detection model

We employ Retina-Net [33] for detecting eyeballs from fMRI images. Retina-Net is one of the most object detection frameworks proposed for 2D natural images. While it can perform well in 3D medical images for object detection. The structure is displayed in Fig. 3. The blue line represents the 3D convolutional layer, and the orange lines represent 3D residual blocks. The 3D feature volumes output from these blocks are considered as $\{C1, C2, C3, C4, C5\}$. $\{M5, M4, M3\}$ are the top-down feature volumes, which are the outputs of feature pyramid networks (FPN). The class subnets and box subnets are denoted as P3-P6, which can classify the anchor boxes and predict the coordinates of the bounding box [34].

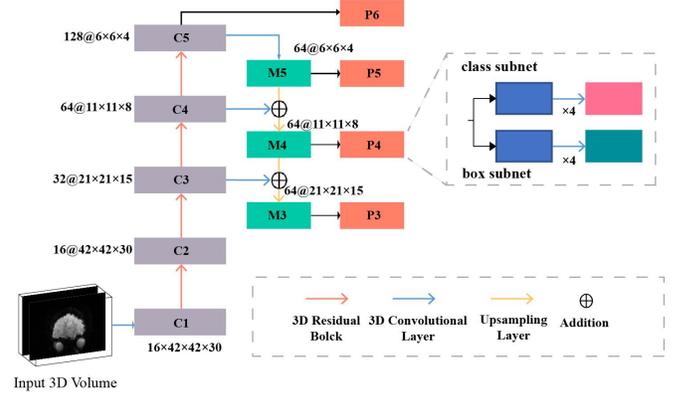

Fig. 3. Illustration of the Eye Detection Model. The orange lines are the 3D residual blocks, the blue lines are the 3D convolutional layers. The yellow lines are the upsampling layers. The backbone of the Retina-Net attaches two subnetworks denoted as P3, P4, P5, and P6, one for classifying anchor boxes and one for regressing bounding boxes. The grey boxes denoted as C1, C2, C3, C4, and C5 are the output feature of the residual blocks. The green boxes denoted as M5, M4, and M3 are the top-down feature volumes corresponding to C5-C3. The P3-P5 represent the class subnets and box subnets.

### D. Performance evaluation

Two Res-Nets are used to acquire visual angles in the x and y directions for eye gaze regression. Choosing the centre of the screen as an original point, the visual angle in this paper is defined as (3):

$$\theta = \arctan\left(\frac{S}{D}\right) \quad (3)$$

Where $\theta = \begin{pmatrix} \theta_H \\ \theta_V \end{pmatrix}$ represents the visual angle in horizontal and vertical directions, $S = \begin{pmatrix} S_x \\ S_y \end{pmatrix}$ represents the horizontal and vertical component distances from the fixation point A to the original point O. Supposing that the eyesight falls on the centre of the screen, D is the distance between the centre of the eye's entrance pupil and the screen. The calculation of the visual angle is shown in Fig. 4 Generally, parameters (e.g., screen size, resolution, or D) in different models of devices are not the same. This causes the absolute coordinate values of the same visual sight to be different. Since the visual angle is not affected by these parameters, it can be used as a common indicator to describe the visual sight, and the trained models can also be easily transferred to the data collected from other labs by using it as training labels.

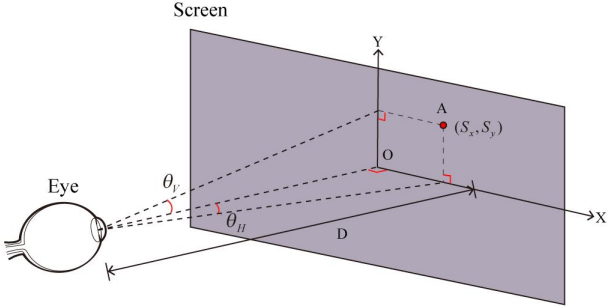

Fig. 4. Illustration of the Calculation of Visual Angle. The grey area is the screen in the experimental environment. A is the fixation point, and O is the centre of the screen. Supposing that the eyesight falls on the centre of the screen, D is the distance between the centre of the eye's entrance pupil and the screen. The visual angle is calculated for vertical and horizontal coordinates. $\theta_V$ represents the vertical coordinate visual angle and $\theta_H$ represents the horizontal coordinate visual angle.

For classification tasks such as outlier detection and saccade tasks, the F1 score, precision score, and recall score are used to evaluate the performance of the models. For regression tasks, in order to consider the performance of the model in both horizontal and vertical directions, Euclidean error (**EE**) serves as performance evaluation (4):

$$EE = \sqrt{\sum_{i=1}^{n}(y_i - \hat{y}_i)^2} \qquad (4)$$

where $y_i$ is the real gaze position and $\hat{y}_i$ is the predicted results; n is equal to 2, which represents the x and y directions. For each volume of each participant, the EE is calculated and then averaged at the individual level. In order to simplify the process, MSE will be regarded as a prediction error (**PE**), which keeps the concept with [16]. The Pearson correlation coefficient (**Pearson's r**) is a statistic for measuring the similarity of temporal trends. The mean absolute error (**MAE**) will also be used to measure the error of vertical and horizontal visual angles.

### E. OUTLIER DETECTION

We noticed that some subjects failed in the eye tracking regression, which had high Euclidean error (EE), and the results even did not express the trend of gaze labels on the time scale. It is emphasized that there are many children in the HBN dataset, and children have high errors in eye movement tasks compared with adults (children with ADHD have higher errors), as seen in [3]. Hence, there is reason to believe that some subjects, especially the younger, did not always fixate on the dot shown on the screen accurately or correctly when they scanned eye movement tasks. In addition, a lower signal-to-noise ratio (SNR) and larger variance of voxels around eyeballs also distort the information about eye gaze. For these reasons, the models cannot decode eye gaze from features.

Frey, Nau and Doeller [16] also mentioned this phenomenon. They regarded these samples that had high PE as outliers and removed them from subsequent analysis. We keep pace with [16] to retain the median performance of 80% of participants for horizontal comparison and exclude the influence of abnormal samples, which improves the reliability of the results. The problem is that the ground truth labels are not available in reality, so we cannot calculate EE or PE to determine if it is an outlier. However, we train a supervised learning model to classify whether it is an abnormal sample at the feature level in that we can remove the bad data before prediction. For the 398 participants of Peer3, the top 15% of participants with high PE were selected and annotated as outliers. The top 15% of participants with lower PE were labelled as normal., which can ensure that the difference between the two types is significant. The results will be shown in the next section.

### IV. EXPERIMENTAL RESULTS

In this part, we begin to discuss a series of experiments to evaluate the performance of the pipeline. The experiments include the performance and transfer capability of models, object detection of eyes model-based naturalistic viewing tasks, pro- and anti-saccade tasks, and the effects of head motion and ages on regression results. Finally, we propose a method to remove outliers before prediction.

#### A. FIXATION TASK REGRESSION
##### 1) ACROSS-SCAN REGRESSION

In this experiment, the models were trained by data from the Peer1 and Peer2 tasks and tested on the Peer3 tasks. Fig. 5(a) shows a sample of eye gaze prediction in the X and Y directions and Fig. 5 (b) shows the heatmap of predictions in the X and Y directions of all the subjects. The predictions are highly correlated with the ground truth. Frey, Nau and Doeller [16] indicated the existence of outliers (data with top 20% PE are considered as outliers), so we also followed the process and removed the participants with high PE (Fig. 5 (c)). After outliers were removed, the proposed model achieved the following: x-direction, MAE=1.11°±0.69°, r=0.91±0.10; y-direction, MAE=1.12°±0.48°, r=0.87±0.11 (Fig. 5 (d), (e)). The performance of the proposed models with Res-Net of 8 layers and 20 layers were also evaluated, and the result shows that there are merely differences among different scales of networks (TABLE 1).

##### 2) ACROSS-INDIVIDUAL REGRESSION

For each subject, there are three Peer tasks (Peer1, Peer2, Peer3), but they do not always contain all of these, as described in the Data Description section. We selected Peer1 of 702 participants, Peer2 of 399 participants, and Peer3 of 398 participants after extracting eye signals from fMRI. The data were split into a training set and a test set. All of the test sets are from Peer1, and the corresponding participants from the test set have no Peer2 and Peer3 scans, which ensures that the models cannot 'see' the subjects when training. In the end, the training set contains 1349 peer tasks, and the testing set contains 150 peer tasks. The results for the across-participant regression are shown in Fig. 6. The performance (20% of participants with high PE are removed) of the models in the testing set: x-direction, MAE=1.70° ± 0.82°, r=0.84 ± 0.12; y-direction, MAE=1.79°±0.67°, r=0.75±0.18 (mean±std). The results verify that Res-Nets can successfully decode eye gaze directions from raw fMRI eyeball signals and that the predictions have low MAE compared with the ground truth when generalized to different subjects.

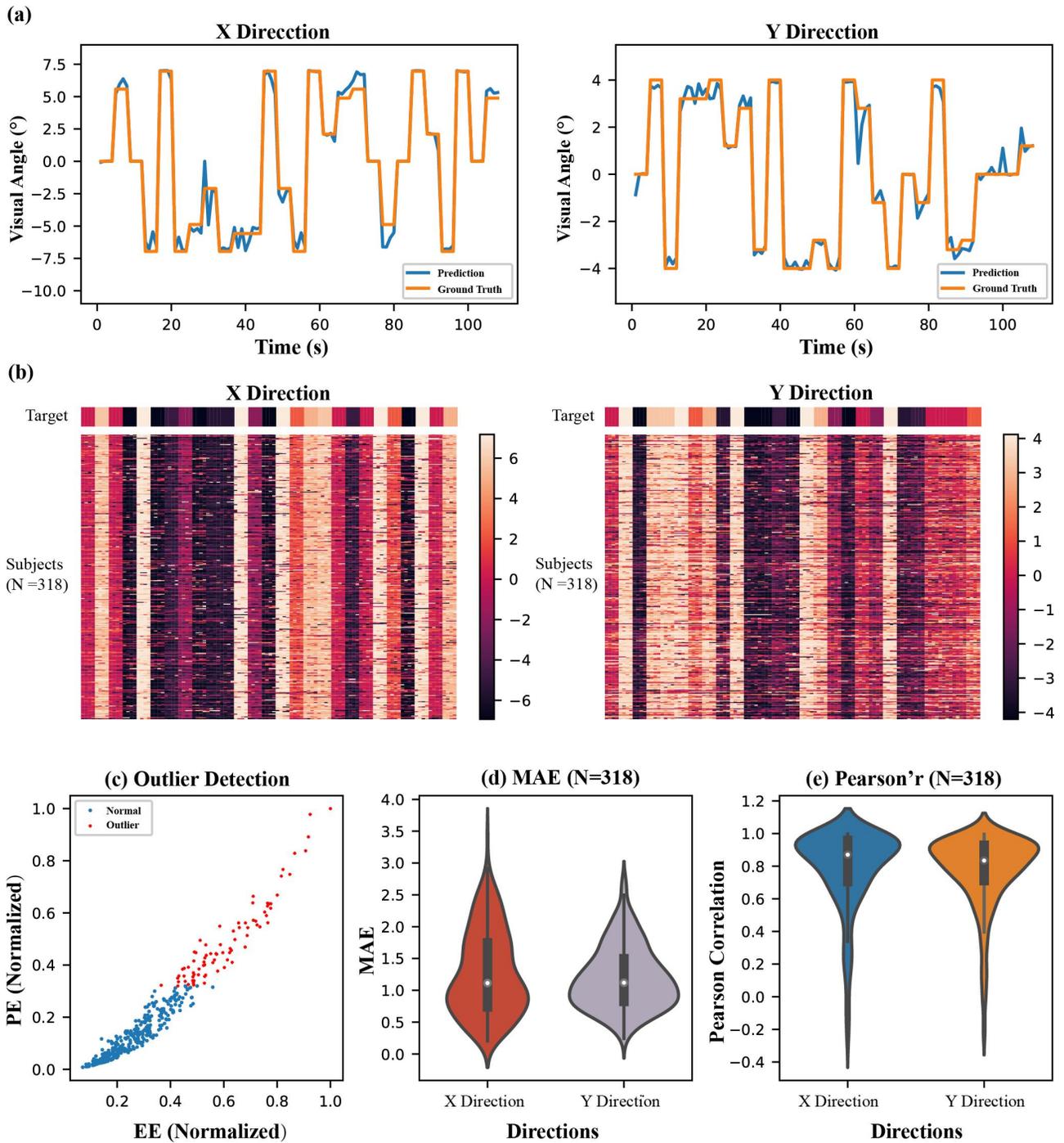

Fig. 5. The results of across-scan regression of the eye fixation task. (a) A sample of eye gaze prediction in the X and Y directions; (b) Heatmap of predictions in the X and Y directions of all the subjects. Each row is the result of a participant in the time series, the Y coordinate is the subjects, and the value of the heatmap is the visual angles predicted by the models. The bar of the top indicates the position of the fixation target; (c) Data with top 20% PE are considered as outliers and removed from further analysis; (d) Violin plots that represent 318 participants' MAE scores with outliers removed (median±median absolute deviation [MAD]) in the X (MAE=1.11°±0.69°, median±MAD) and Y (MAE=1.12°±0.48°, median±MAD) directions; (e) Violin plot that represents 318 participants' the Pearson correlation coefficient (with outliers removed) in the X (r=0.91±0.10, median±MAD) and Y (r=0.87±0.11, median±MAD) directions.

TABLE 1 THE REGRESSION MODEL PERFORMANCE IN THE ACROSS SCANS TASKS

| Method | EE (outlier removal) | Mean Pearson's r (outlier removal) | Pearson's r (median ± MAD) | | MAE (median ± MAD) | |
|---|---|---|---|---|---|---|
| | | | X direction | Y direction | X direction | Y direction |
| J. Son [11] | | | 0.78±0.21 | 0.68±0.29 | 1.97°±1.14° | 1.48°±0.75° |
| M. Frey [16] | 2.89° | 0.86 | | | | |
| Res-Net 8 | 2.19° | 0.86 | 0.86±1.04 | 0.81±0.68 | 1.58°±1.04° | 1.37°±0.68° |
| Res-Net 12 | **2.04°** | **0.87** | **0.87±1.03** | **0.83±0.65** | **1.42°±1.03°** | **1.26°±0.65°** |
| Res-Net 20 | **2.04°** | 0.86 | 0.86±1.03 | 0.83±0.65 | 1.45°±1.03° | 1.27°±0.65° |

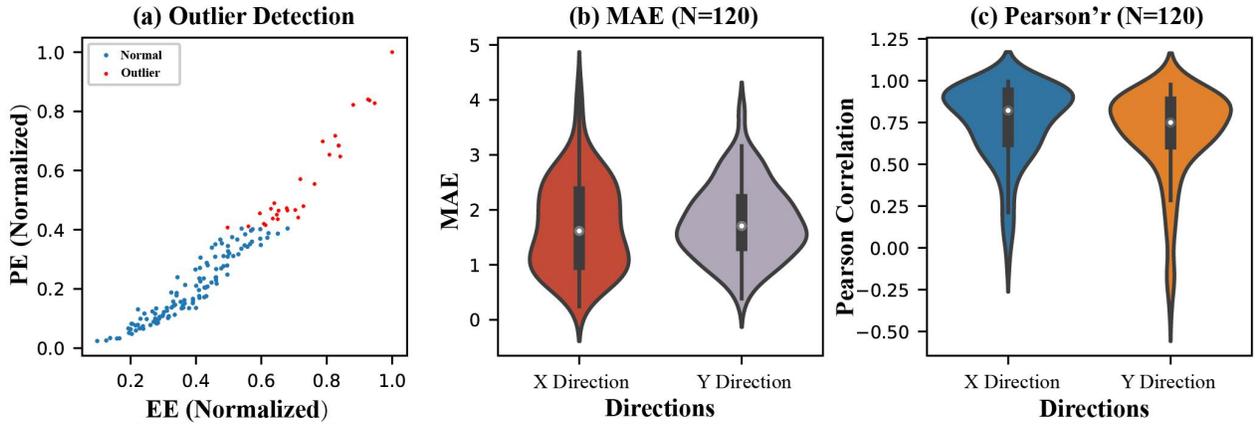

Fig. 6. The results of across-individual regression of the eye fixation task. (a) Data with top 20% PE are considered as outliers and removed from further analysis; (b) Violin plots that present the 120 participants' MAE scores with outliers removed in the X(MAE=1.70°±0.82°, mean±std) and Y (MAE=1.79°±0.67°, mean±std) directions; (c) Violin plot that presents the 120 participants' the Pearson correlation coefficient with outliers removed (median±median absolute deviation [MAD]) in the X (r=0.84±0.12, mean±std) and Y (r=0.75±0.18, mean±std) directions.

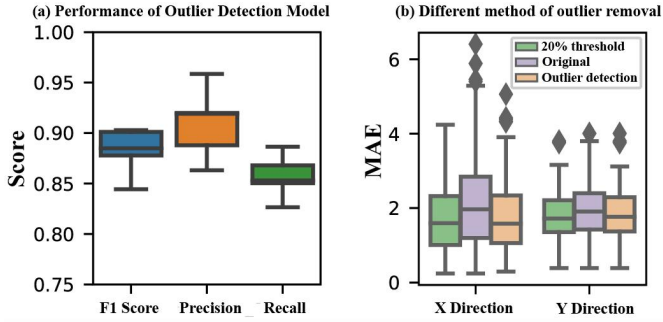

Fig. 7. The results of the outlier detection model. (a) The performance of the outlier detection model. F1= 88.22%±2.1, Precision=90.97%±3.23, Recall=85.68%±2.00. (b) The EE of data with 20% high PE removed, the original samples, and removal by the outlier detection model in x and y directions. The outlier detection model trained by Peer3 can remove the outlier samples and make the performance comparative to the 20% high PE thresholding.

### 3) THE END-TO-END PIPELINE FOR EYE GAZE REGRESSION

Although the digital morphological algorithm with specific parameters (the shape of structural elements, the value of binary threshold) is able to extract eye signals from most of the raw fMRI images, there are still some participants who fail to get eyeballs and need to adjust the parameters manually. To avoid artificial involvement, we trained a 3D object detection network to grasp eyeballs from fMRI, which concatenates the regression networks (R) to implement an end-to-end pipeline for eye gaze regression. We used the morphology algorithm to obtain the signals and the bounding boxes of eyes in the naturalistic viewing data that the bounding boxes have been checked and ensured to contain the whole eye signals. Therefore, the fMRI and bounding box can be used to train a 3D retina net (the details about the structure have been described in the Deep Learning Structure section) to detect eyeballs. Then, the trained 3D retina-net can detect eyeballs in eye gaze tasks. The across-individual tasks were reproduced, but the morphology algorithm (M) was replaced by retina-net (DT), and the removal of the top 20% of samples with high PE (20% PE threshold) was replaced by an outlier detection model (O) before regression to verify the feasibility of the pipeline (TABLE 2). The morphology algorithm costs 0.03 s for each volume; the object detection model costs 0.01 s for each volume; and the predicted model costs 0.005 s for each volume.

### B. NATURALISTIC VIEWING REGRESSION

In reality, naturalistic viewing has become a popular stimulus to exploit eye movement behaviours related to brain activity. In the HBN dataset, some of the subjects are asked to watch movies during scans according to descriptions of imaging protocols. We select one of the movie tasks, 'the Present' (TP), to test the performance of our pipeline. Before regression, 214 participants remained, which had complete eye signals and no visible movementartifacts. In EEG protocols, 1812 participants were collected with eye tracker data, which they watched 'the Present' too. 800 ms windows were used to match the sampling rate (TR=800 ms) of the MR scans and for each of the windows, the median of the raw sample was regarded as the eye gaze direction recorded by the eye tracker. Therefore, 250 time points were extracted from the eye-tracking data. There were some participants who missed some of the 250 eye tracking time points because of eye blinking or time off the screen. Thus, the participants with 15% of missing eye tracking time points were discarded, and 1351 participants of eye tracking time points were averaged, converted to visual angle, and normalized. The initial parameters of the models were trained in the across scans task, and the predictions of 214 participants were averaged and normalized. Finally, the pipeline achieves the following: x direction, r=0.88; y direction, r=0.84. The results stand with [4], [35], [36] that movies evoke highly consistent and reproducible eye movement across different observers (Fig. 8).

TABLE 2 THE PERFORMANCE OF THE END-TO-END PIPELINE ACROSS INDIVIDUAL TASKS

| Method | EE (mean ± std) | Mean Pearson's r (mean ± std) | Pearson's r (mean ± std) X direction | Y direction | MAE (mean ± std) X direction | Y direction |
|---|---|---|---|---|---|---|
| M+R | 3.43°±1.43° | 0.71±0.23 | 0.76±0.23 | 0.67±0.27 | 2.21°±1.32° | 1.96°±0.74° |
| DT+R | 4.20°±1.56° | 0.60±0.26 | 0.65±0.28 | 0.55±0.29 | 2.94°±1.43° | 2.23°±0.73° |
| M+R+20% PE threshold | 2.89°±0.94° | 0.80±0.13 | 0.84±0.12 | 0.75±0.18 | 1.70°±0.82° | 1.79°±0.67° |
| DT+R+20% PE threshold | 3.62°±1.09° | 0.70±0.16 | 0.75±0.18 | 0.64±0.21 | 2.41°±0.99° | 2.04°±0.64° |
| M+O+R | 3.03°±1.13° | 0.78±0.16 | 0.83±0.16 | 0.73±0.22 | 1.81°±1.00° | 1.85°±0.73° |
| DT+O+R | 3.42°±1.17° | 0.76±0.13 | 0.80±0.14 | 0.71±0.17 | 2.24°±1.02° | 1.94°±0.70° |

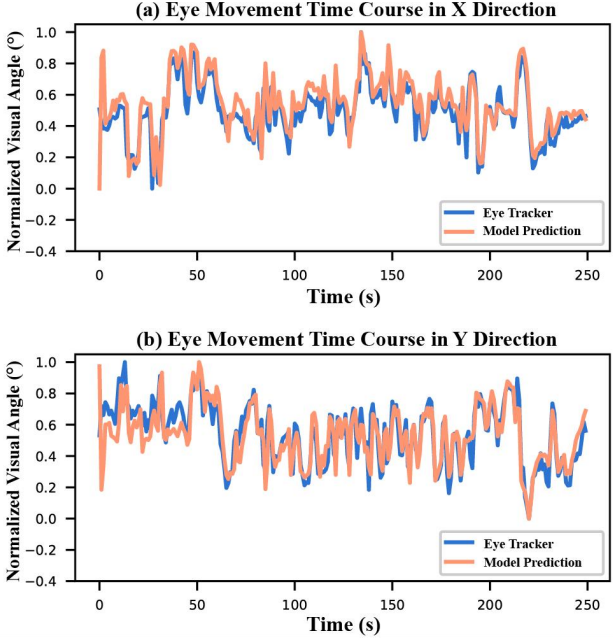

Fig. 8. The regression of the naturalistic viewing task. (a) The averaged eye movement time course in x direction (r=0.88). (b) The averaged eye movement time course in y direction (r=0.84). The orange line is the prediction of the Res-Net model, and the blue line is the mean eye-tracking data.

### C. ACROSS-DATASET LEARNING OF PRO-SACCADE AND ANTI-SACCADE TASKS

The pro-saccade tasks of data in the 'visual guide saccade' were used for the training model. The parameters of the visual angle calculation are not available, so the absolute error of the pixel distance between the model prediction and ground truth has to be calculated for evaluation.

Fivefold validation was used for training, and the results in each fold are 165.2±146.9(1st fold), 125.2±105.0(2nd fold), 132.6±116.8(3rd fold), 114.8±115.6(4th fold), 135.0±112.5 (5th fold). (Fig. 9(a)). Subsequently, the model with the best result of the fivefold validation in the pro-saccade task is determined as the prediction model to be used in the anti-saccade task. For the anti-saccade task, although the real coordinates of the subject's gaze on the periphery cannot be obtained, this task focuses more on whether the subject has completed the process of staring in the opposite direction of the stimulus target. Hence, the samples from the anti-saccade task labeled 'correct' are fed into the model for prediction. If the output result is in the opposite direction of the stimulus target of the anti-saccade task, the model is considered to have successfully decoded the anti-saccade behavior from the fMRI signal. Finally, our model achieves precision=89.2%, recall=87.9%, and F1 score =88.5% in the 'visual guide saccade' and precision=88.4%, recall=67.5%, and F1 score =76.5% in the 'intensive anti-saccade'. The confusion matrixes of the anti-saccade task are shown in Fig. 9(b).

By showing the results, it is concluded that: 1. The model can decode the gaze information of the eyeball even with few samples. 2. The model trained with few samples can be used in qualitative eye-tracking tasks such as anti-saccade. 3. The model achieves good generalization ability on tasks across datasets.

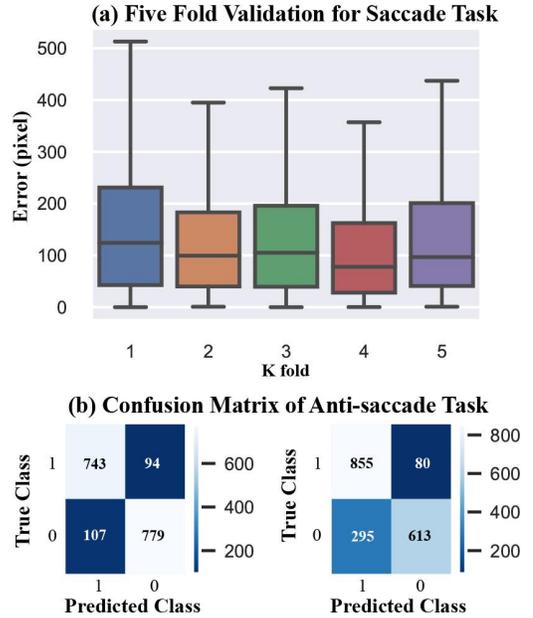

Fig. 9. The performance of fivefold validation in pro-saccade tasks. (a) The errors in each fold: 165.2±146.9, 125.2±105.0, 132.6±116.8, 114.8±115.6, 135.0±112.5. (b) The confusion matrix of the anti-saccade task.

### D. IMPACT ABOUT HEAD AND AGES

Head motion and ages have also become important factors impacting regression results (not removing outliers). The Pearson coefficient (Normalized) of all samples is used as a measure of the extent of regression, and mean FD Jenkinson is used as a measure of head motion. Logistic regression revealed that age was positively correlated with regression accuracy (beta=0.14, P value=1.04 × 10-29) and that head motion was negatively correlated with regression accuracy (beta=6.02, P value=4.25 × 10-17). The results are described in Fig. 10. Although these factors significantly affect the regression results, the effect is not enough to cause the regression to fail (exhibiting outliers).

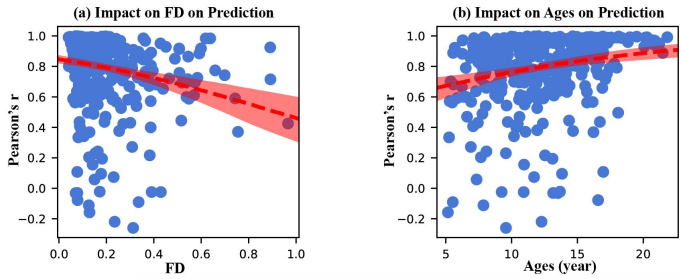

Fig. 10. The impact of age and head motion on prediction performance. The red line is the logistic regression result, and the light red area presents a 95% confidence interval of regression. (a) The head motion indices, FD, showed a negative correlation with the accuracy (beta=6.02, P value=4.25×10-17); (b) The age showed a positive correlation with the regression accuracy (beta=0.14, P value=1.04×10-29).

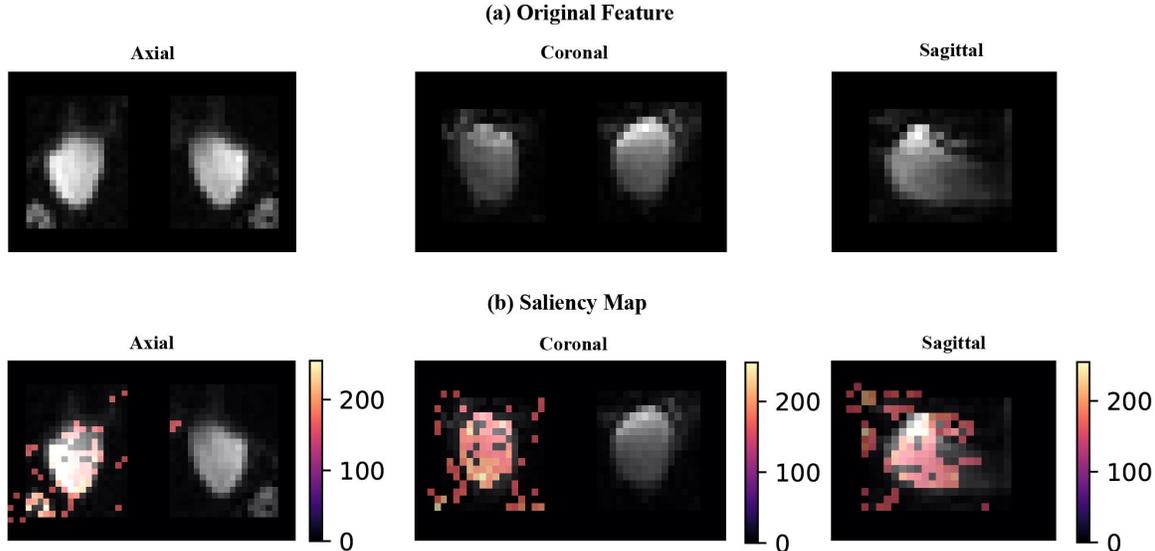

Fig. 11. The saliency map by the guided Grad-CAM method. The left, middle and right columns represent the features (eyeballs) in coronal, vertical and sagittal sections. (a) Raw fMRI eye signals. (b) The saliency map generated by the guided Grad-CAM method covering the original image. It shows that the model tends to pay more attention on one eye information for regression.

### E. WHAT INFORMATION DOES THE MODEL RELY ON TO ACHIEVE REGRESSION?

To explain the question, [37] proposed a method to highlight the important region of the feature level for prediction, which can help to directly realize which part of the information the networks prefer. The results are shown in Fig. 11. Because most of the areas of features have low salience values, a threshold is chosen to filter out the low value and retain the high salience area. We can conclude that: 1. Most of the information is concentrated on eyeballs. This finding supports the conclusion of [8], [9], [11] that eye signals provide adequate information for eye fixation. Some medical image segmentation algorithms, such as U-Net, have become more suitable for extracting eyeballs to reduce redundant information caused by noise and nerve tissue around the eyes. 2. Res-Net tends to use information from one eyeball for regression. (Another eye has a lower salient value and is not shown in the figure.) What the conclusion gives us is that the data augmentation can be achieved by mirroring flipping eyeball images.

## V. CONCLUSIONS

In this paper, we propose a deep learning-based pipeline to predict eye gaze information for fMRI experiments and achieve camera-less regression from raw fMRI images, which leads to time and cost savings in the entire experimental process. The work consists of mathematical morphological operations for eye signal extraction and Res-Net models for eye gaze regression. Mathematical morphology helps to extend the scale use of data and avoids intricate preprocessing of fMRI. Deep learning produces robust and precise results for sequential noncamera eye tracking. Moreover, a series of experiments designed according to cognitive eye movement tasks evaluate the reliability and practicability of the pipeline. Hence, the work can be employed during fMRI scans relative to eye movement in psychological and cognitive research.

Screen setups vary across MRI scanners but always remain the same within one MRI scanner. Our cross-dataset result validated MRGazer's generality to different research within the same MR scanner. Thus, a recommended way to apply MRGazer to a new MR scanner is to transfer learning with a few fixations and saccade fMRI data collected in the scanner and apply it to the remaining fMRI data from the same scanner. Future work including information on the field of view (FOV), head shape, head position, interpupillary distance, and eyeball size may improve the model's flexibility and generalization capability. Moreover, in addition to fixation, eye movements include saccade, blink, pursuit, etc. A futural model decoding all these eye movements at higher temporal resolution (tens of milliseconds) may be helpful in MR-based eye tracking research

### ACKNOWLEDGMENT

This work was partially supported by National Natural Science Foundation of China (81701665), The University Synergy Innovation Program of Anhui Province (GXXT-2021-003).